\documentclass[11pt]{article}
\usepackage[utf8]{inputenc}
\pdfoutput=1
\usepackage[sc,osf]{mathpazo}
\usepackage{booktabs}       
\usepackage[height=8.5in,width=6in,letterpaper]{geometry}
\usepackage[sort&compress]{natbib}
\usepackage[colorlinks=true,citecolor=blue,breaklinks]{hyperref}
\usepackage[hyphenbreaks]{breakurl} 
\usepackage[tracking]{microtype}

\usepackage{enumerate}
\usepackage{enumitem}   

\usepackage[pdftex]{graphicx}

\usepackage{bbm}          

\usepackage{amssymb,amsmath,amsthm, mathtools}
\usepackage{amscd}

\usepackage{caption, subcaption}
\usepackage{float} 

\usepackage{fancyhdr}
\usepackage{titlesec}

\usepackage{algorithm}
\usepackage{algorithmic}

\makeatletter
\newlength\aftertitskip     \newlength\beforetitskip
\newlength\interauthorskip  \newlength\aftermaketitskip

\renewcommand\thefootnote{\textcolor{red}{\arabic{footnote}}}

\def\maketitle{\par
 \begingroup
   \def\thefootnote{\color{red}\fnsymbol{footnote}}
   \def\@makefnmark{\hbox to 4pt{$^{\@thefnmark}$\hss}}
   \@maketitle \@thanks
 \endgroup
\setcounter{footnote}{0}
 \let\maketitle\relax \let\@maketitle\relax
 \gdef\@thanks{}\gdef\@author{}\gdef\@title{}\let\thanks\relax}

\def\@startauthor{\noindent \normalsize\bf}
\def\@endauthor{}
\def\@starteditor{\noindent \small {\bf Editor:~}}
\def\@endeditor{\normalsize}
\def\@maketitle{\vbox{\hsize\textwidth
 \linewidth\hsize \vskip \beforetitskip
 {\begin{center} \LARGE\@title \par \end{center}} \vskip \aftertitskip
 {\def\and{\unskip\enspace{\rm and}\enspace}%
  \def\addr{\small\it}%
  \def\email{\hfill\small\tt}%
  \def\name{\normalsize\bf}%
  \def\AND{\@endauthor\rm\hss \vskip \interauthorskip \@startauthor}
  \@startauthor \@author \@endauthor}
}}

\makeatother

\pdfoutput=1                    

\usepackage[export]{adjustbox}

\usepackage{etoolbox}
\makeatletter
\patchcmd{\nocite}{\ifx\@onlypreamble\document}{\iftrue}{}{}
\usepackage{todonotes}

\def\R{\mathbb{R}}

\def\K{\mathbb{K}}

\def\cX{\mathcal{X}}
\def\cY{\mathcal{Y}}
\def\cZ{\mathcal{Z}}

\def\cB{\mathcal{B}}
\def\cM{\mathcal{M}}

\def\cV{\mathcal{V}}

\newtheorem{theorem}{Theorem}[section]

\newtheorem{lemma}[theorem]{Lemma}
\newtheorem{lemma*}{Lemma}

\def\suchthat{\medspace|\medspace}

\def\diag{\text{diag}}

\DeclareMathOperator*{\argmin}{argmin}
\DeclareMathOperator*{\argmax}{argmax}
\DeclarePairedDelimiterX{\infdivx}[2]{(}{)}{%
  #1\;\delimsize\|\;#2%
}
\newcommand{\KL}{\text{KL}\infdivx}

\graphicspath{{figs/}}

\title{Structured Optimal Transport}

\author{\name David Alvarez-Melis \email{dalvmel@mit.edu}\\
  \name Tommi S. Jaakkola \email{tommi@csail.mit.edu}\\
  \name Stefanie Jegelka \email{stefje@csail.mit.edu}\\
  \addr{Massachusetts Institute of Technology}
}

\begin{document}
\maketitle

\begin{abstract}
  Optimal Transport has recently gained interest in machine learning for applications ranging from domain adaptation, sentence similarities to deep learning. Yet, its ability to capture frequently occurring structure beyond the ``ground metric'' is limited. In this work, we develop a nonlinear generalization of (discrete) optimal transport that is able to reflect much additional structure. We demonstrate how to leverage the geometry of this new model for fast algorithms, and explore connections and properties.  Illustrative experiments highlight the benefit of the induced structured couplings for tasks in domain adaptation and natural language processing.
\end{abstract}


\section{Introduction}

Optimal transport provides a natural, elegant framework for comparing probability distributions while respecting the underlying geometry \citep{villani2008optimal}. Due to its strong theoretical foundations and many desirable properties, both the continuous and discrete versions of the transportation problem have received considerable attention in various fields within and beyond mathematics, including statistics \citep{Mallows1972Asymptotic}, differential equations \citep{Jordan1998Variational}, optics \citep{Glimm2003Optical} and economics \citep{galichon2016optimal}. Within machine learning and related fields, optimal transport distances (in particular the Wasserstein metric) have found successful application to shape analysis \citep{Gangbo2000Shape}, image registration and interpolation \citep{Solomon2015Convolutional}, domain adaptation \citep{Courty2017Optimal}, adversarial neural networks \citep{Arjovsky2017Wasserstein}, and multi-label prediction \citep{Frogner2015Learning}. The discrete version of the problem has also had impact in settings where relaxed notions of matchings are sought, such as pairing control and treatment units in observational studies \citep{rosenbaum1985constructing}. The range of applications has been growing with the development of fast algorithms \citep{cuturi2013sinkhorn, genevay2016stochastic}.

An important appeal of optimal transport distances is that they reflect the metric of the underlying space in the transport cost. Yet, in a number of settings, there is further important structure that remains uncaptured. This structure can be \emph{intrinsic} if the distributions correspond to structured objects (e.g., images with segments, or sequences) or \emph{extrinsic} if there is side information that induces structure (e.g.~groupings). A concrete example arises when applying optimal transport to domain adaptation, where a subset of the source points to be matched have known class labels. In this case, we may desire source points with the same label to be matched coherently to the same compact region of the target space, preserving compact classes, and not be split into disjoint, far locations \citep{Courty2017Optimal}. When pairing control and treatment units in observational studies of treatment effects, it is beneficial to compare treated and control subjects from the same ``natural block'' (e.g., family, hospital) so as to minimize the difference between unmeasured covariates \citep{pimentel2015large}. In all these examples, the additional structure essentially seeks correlations in the mappings of ``similar'' source points. Such dependencies, however, cannot be induced by standard formulations of optimal transport whose cost is separable in the mapping variables\footnote{The original optimal transport formulation with cost $\sum_{ij}c_{ij}\gamma_{ij}$ is linear in the mappings $\gamma_{ij}$, $\gamma_{kl}$ of separate source locations $i$, $k$; the mappings are counted independently.}; they require nonlinearities.

In this work, we develop a framework to incorporate such structural information directly into the optimal transport problem. This novel formulation opens avenues to a much richer class of (nonlinear) cost functions, allowing us to encode known or desired interactions of mappings, such as grouping constraints, correlations, and explicitly modeling topological information that is present, for instance, in sequences and graphs. The tractability of this nonlinear formulation arises from polytopes induced by submodular set functions. Submodular functions possess two highly desirable properties for our problem: (1) they naturally encode combinatorial structure, via diminishing returns and as combinatorial rank functions; and (2) their geometry leads to efficient algorithms.

The resulting combination of the geometries of transportation and submodularity leads to a problem with rich, favorable polyhedral structure and connections to game theory and saddle point optimization. We leverage this structure to solve the \emph{submodular optimal transport} problem via a saddle-point mirror prox algorithm involving alternating projections onto the probabilistic polytope defined by the transportation constraints and the base polytope associated with the submodular cost function. The former can be done efficiently through Sinkhorn iterations, while the latter, as we will see, can be solved exactly in $O(n \log n)$ time for a suitable class of submodular functions.

Via various applications and experiments, we explore the characteristics of the solutions to this novel transportation problem and demonstrate the efficiency of our algorithms.
We show how different submodular functions yield solutions that interpolate between strictly structure-aware transportation plans and structure-agnostic regularized versions of the problem.  Besides these synthetic experiments, we evaluate our framework in two real-life applications: domain adaption for digit classification and sentence similarity prediction. In both cases, introducing structure leads to better empirical results.

\paragraph{Contributions.} In short, we make the following contributions: (1) we propose a framework for including structured information into optimal transport that integrates concepts from combinatorics to geometry; (2) we show efficient optimization methods that carefully exploit the geometric structure; (3) we demonstrate the utility of this new framework via example applications in domain adaptation and sentence similarity, where our structured couplings outperform classical and class-regularized versions of optimal transport.


\section{Background}

\subsection{Optimal Transport}

The original formulation of optimal transport by Gaspar Monge considers two probability measures $\mu,\nu$ over metric spaces $\cX,\cY$, and a measurable cost function $c: \cX \times \cY \rightarrow \R$, which represents the cost of transporting a unit of mass from $x \in \cX$ to $y \in \cY$. The problem asks to find a transport map $T: \cX \rightarrow \cY$ that realizes
\begin{equation}\label{monge_ot}
	\inf_{T} \left\{ \int_{\cX}c(x,T(x))d\mu(x) \suchthat T_{\#}\mu = \nu \right\},
\end{equation}
where $T_{\#}$ denotes the push-forward of $\mu$ by $T$. The solution to \eqref{monge_ot} might not exist. However, a convex relaxation of the problem due to Kantorovich is guaranteed to have a solution:
\begin{equation}\label{kantor_ot}
	\inf_{\gamma} \left\{ \int_{\cX \times \cY}c(x,y)d\gamma(x,y) \suchthat \gamma \in \Gamma(\mu,\nu) \right \},
\end{equation}
where $\Gamma(\mu,\nu)$ is the set of \emph{transportation plans}, i.e., joint distributions with marginals $\mu$ and $\nu$.
If $\mu$ and $\nu$ are only available through discrete samples $\{\mathbf{x}^s_i\}_{i=1}^n$, $\{\mathbf{x}^t_i\}_{i=1}^m$, the empirical distributions can be written as
\begin{equation}
	\mu = \sum_{i=1}^n p_i^s\delta_{\mathbf{x}_i^s}, \quad \nu = \sum_{i=1}^m p_i^t\delta_{\mathbf{x}_i^t}
\end{equation}
where $p_i^s,p_i^t$ are the probabilities associated with the samples. It is easy to adapt Kantorovich's formulation to this discrete setting. In this case, the space of transportation plans is a polytope:
\begin{equation}\label{transport_poly}
	\cM_{\mu,\nu} = \{ \gamma \in \mathbb{R}^{n \times m}_+ \suchthat \gamma \mathbf{1} = \mu, \medspace \gamma^T\mathbf{1} = \nu \}
\end{equation}
The cost function only needs to be specified for every pair $(\mathbf{x}^s_i, \mathbf{x}^t_j)$, i.e., it is a matrix $C \in \R^{n \times m}$, and the total transportation cost incurred by $\gamma$ is $\sum_{ij} \gamma_{ij} c_{ij}$. Thus, the discrete optimal transport (DOT) problem consists of finding a transport plan that solves

\begin{equation}\label{eq:original_OT}
	\min_{\gamma \in \cM_{\mu,\nu}} \langle \gamma, C \rangle .
\end{equation}
If $n=m$, and $\mu$ and $\nu$ are uniform measures, $\cM_{\mu,\nu}$ is the Birkhoff polytope of size $n$, and the solutions of \eqref{eq:original_OT}, which lie in the corners of this polytope, are permutation matrices.

Discrete optimal transport is a linear program, and thus can be solved exactly in $O(n^3 \log n)$ with interior point methods. In practice, a version with entropic smoothing has proven more efficient \citep{cuturi2013sinkhorn}:
\begin{equation}
\min_{\gamma \in \cM}\; \langle \gamma, C \rangle - \frac{1}{\lambda}H(\gamma).
\end{equation}
The solution of this strictly convex optimization problem has the form $\gamma^* = \diag(u)\K\diag(u)$, with $\K = e^{-\frac{C}{\lambda}}$ (entrywise), and can be obtained fast via the Sinkhorn-Knopp algorithm, an iterative matrix-scaling algorithm \citep{cuturi2013sinkhorn}. Besides significant speedups, the smoothed problem often leads to better empirical results in downstream applications.

\subsection{Submodularity}
A set function $F:2^V \rightarrow \R$ over a ground set $V$ of items is called \emph{submodular} if it satisfies \emph{diminishing returns}: for all $S \subseteq T \subseteq V $ and all $v$ in $ V \setminus T$, it holds that
\begin{equation}\label{def:submod}
	F(S \cup \{v\})  - F(S) \geq F(T \cup \{v\}) - F(T)
\end{equation}
$F$ is called supermodular if $-F$ is submodular, and modular if it is both sub- and supermodular. The tractability of submodular functions arises from the polytopes they define. The \emph{base polytope} of $F$ is
\[ \mathcal{B}_F = \{ y \in \mathbb{R}^{|V|} \suchthat  y(\cV) = F(\cV);\; y(A) \leq F(A)\; \forall A \subseteq \mathcal{V} \}. \]
Base polytopes generalize matroid polytopes (convex hulls of combinatorial ``independent sets''), and lead to strong links with convexity.

The \emph{Lov\'asz extension} of a set function $F$ extends its domain from $2^V$ to $\R^n_{+}$ \citep{lovasz83mathematical}. For any $w \in \R^n_{+}$, order its coordinates so that $w_1 \geq \dots \geq w_n$ and define $w_{n+1} = 0$ and $S_j = \{ i \suchthat w_i \geq w_j \}$. The Lov\'asz extension $f$ of $F$ is 
\begin{equation}
	f(w) = \sum_{i=1}^n(w_j - w_{j+1}) F(S_j).
\end{equation}
If $F$ is submodular, the Lov\'asz extension is equivalent to the support function
\begin{equation}\label{def:lovasz}
	f(w) = \max_{x \in \cB_F} w^Tx,
\end{equation}
which is convex. In fact, 
$f$ is convex if and only if $F$ is submodular \citep{lovasz83mathematical}. 

\section{Optimal Transport with Submodular Costs}

In the classical formulation of optimal transport \eqref{eq:original_OT}, the cost function $\langle \gamma, C\rangle$ is linear in the decision variables $\gamma$. This means each potential pairwise assignment $\gamma_{ij}$ (i.e.~every pair $(\mu_i,\nu_j)$) is treated independently. But, in some applications, it is desirable to bias certain points to be mapped \emph{together}, i.e., to introduce dependencies between assignments. 

In our running example of domain adaptation, we want points from the same class to be transported ``together''. Intuitively, the joint cost of mapping points from the same class to close-by target points should be lower than splitting them apart, even if the transportation distances are the same.

More generally, we might want to encourage mappings of subspaces to subspaces, or, on the contrary, discourage some combinations of assignments. A flexible framework to express such interactions over discrete choices is via submodular functions \citep{lin11optimal,Jegelka2011Submodularity, Kohli2013Principled}: intuitively, property \eqref{def:submod} implies that the marginal cost of an additional element decreases as more ``compatible'' items have already been chosen, and thus it is relatively \emph{cheaper} to select compatible items together (e.g., items from the same group) than non-compatible ones. 

To see how submodularity can be leveraged for optimal transport, consider for a moment Monge's formulation \eqref{monge_ot}, where we seek a matching of variables in $U$ and $V$ with minimal cost. Any matching can be expressed as a set of edges $S = \{ (u_1,v_1) , \dots, (u_k,v_k)\}$, and its cost as a set function $F: 2^{|U|\times|V|} \rightarrow \R^+$. Under this formulation, the classic definition of optimal transport uses a \emph{modular} cost function:
\[ F(S) = \sum_{(u,v) \in S} c_{uv} \]
so the cost of an additional match $(u,v)$ is the same, $c_{uv}$, regardless of what assignments have already been made. If we let $F$ be submodular instead, property \eqref{def:submod} implies that the marginal cost of additional edges decreases as the set of matches grows. The magnitude of decrease depends on the sets $A \subseteq B$ and the new item, and the choice of $F$. We will channel this decrease to occur only when the additional ``item'' (assignment $(u,v)$) is compatible with already chosen ``items''. 

\subsection{Submodular cost functions}\label{sec:submod_funs}
The rich class of submodular functions allows various types of structural information (compatibility) to be encoded in the cost function.
As an example, recall the local consistency structure induced by class labels in domain adaptation. We may divide the support of the source and target distributions $\mu$ and $\nu$ into regions (subsets of samples) $U_k \subset U$ and $V_l \subset V$. These induce a partition of the set of assignments too:
\[ E_{kl}:= \{ (u,v) \suchthat u \in U_k, \medspace v \in V_l  \} \]
Now define
\begin{equation}\label{fun:sum_concave}
	F(S) := \sum_{kl}F_{kl}(S \cap E_{kl}),
\end{equation}
where each $F_{kl}$ is submodular with reduced support $E_{kl}$. One possible choice for $F_{kl}$ is
\begin{align}\label{eq:cluster_funs}
	F_{kl} (S)= g_{kl}\biggl ( \sum_{(u,v) \in S \cap E_{kl}}  C_{uv} \biggr),
\end{align}
where $C_{ij} \in \R^+$ is the ground metric cost between $x^s_i$ and $x^t_j$, and 
$g_{kl}: \R \rightarrow \R$ are scalar monotone increasing concave functions whose effect is to dampen the cost of additional edges between the partitions $U_l$ and $V_k$, thus encouraging edge selections that map most of the mass in $U_l$ to the same $V_k$.
To grant discounts only after a sufficient number of assignments have been chosen from a group, we may use an explicit threshold, e.g.,
\begin{equation}\label{fun:thresh_sqrt}
	g_{kl}(x) = \min\{x, \alpha\} + \sqrt{[x-\alpha]_+}.
\end{equation}
We use such functions in the clustered point matching, domain adaptation and sentence similarity experiments in Section~\ref{sec:expts}. We may also use subspaces for encoding structure. For example, a smoother grouping of assignments $(u,v)$ could be encoded by stacking feature vectors for $u$ and $v$ into one vector $\phi(u,v)$ and taking $F(S) = \mathrm{rank}(\Phi_S)$ to be the rank of the matrix of features of the selected assignments, or the volume $F(S) = \log\det(\Phi_S^\top \Phi_S)$. This function captures discrete groups if the feature vectors are indicator vectors of groups. Other examples include hierarchical structures and coverage functions.

\subsection{Problem Formulation: Submodular optimal transport}

The functions defined above have discrete domains, i.e., they correspond to discrete matchings, but we really seek a formulation like \eqref{eq:original_OT}, with continuous, fractional assignments. The key to obtaining a nonlinear, structured analog of Kantorovich's formulation \eqref{kantor_ot} of the classical problem is the convex \emph{Lov\'asz extension} $f$ of the submodular function $F$. The above intuitions and effects carry over, and we define the \emph{submodular optimal transport} problem as
\begin{equation}\label{PolytopeForm}
  \min_{\gamma \in \cM} f(\gamma)\;\; \equiv\;\; \min_{\gamma \in \cM} \max_{\kappa \in \cB_F} \langle \gamma, \kappa \rangle.
\end{equation}
The right hand side follows since the Lov\'asz extension is also the support function of the submodular base polytope. This relaxation has another advantage: while the discrete version is hard to even solve approximately \citep{goel2009approximability}, problem \eqref{PolytopeForm} is a convex optimization problem. 

The new structured optimal transport problem recovers many desirable properties of the original optimal transport. For example, the ``distance'' implied by it is a semi-metric under mild assumptions (proof in the Appendix): 
\begin{lemma}\label{lemma:semi-metric}
	Suppose the ground cost $C(\cdot, \cdot)$ is a metric and that $F$ is a submodular non-decreasing function such that $F(\emptyset)=0$ and $F(\{(i,j)\}) >0 $ iff $C(x_i,y_j)>0$. Then $d_F(\mu,\nu) = \min_{\gamma \in \cM}f(\gamma)$ is a semi-metric.
\end{lemma}

Problem \eqref{PolytopeForm} suggests two possible approaches for computing the optimal transport plan $\gamma^*$. The left-hand side is a non-smooth but convex optimization problem, which can be solved via subgradient methods. Alternatively, the minimax form is a \emph{smooth} convex-concave optimization over nonempty, closed and convex sets.\footnote{$\cM, \cB_F$, being polytopes, are closed and convex. $\cM$ is always nonempty ($\mu\nu^T \in \cM$), and so is $\cB_F$ \citep{bach2013learning}.} Therefore, \eqref{PolytopeForm} is a convex-concave saddle-point problem \citep{Juditsky2011First}. The solutions $z^*:=(\gamma^*, \kappa^*)$ of this problem, i.e., the \emph{saddle points} $\phi:=\langle \cdot, \cdot \rangle$ in $\cZ:=\cM \times \cB_F$, satisfy 
\begin{equation*}\label{eq:saddle_points}
	\phi(\gamma^*,\kappa) \leq \phi(\gamma^*,\kappa^*) \leq \phi(\gamma,\kappa^*) \quad \forall \gamma \in \cM, \kappa \in \cB_F
\end{equation*}
This formulation gives rise to a primal-dual pair of convex optimization problems:
\begin{align}
	 \text{Opt}(P) = \min_{\gamma \in \cM} \bar{\phi}(\gamma), \quad \overline{\phi}(\gamma):= \sup_{\kappa \in \cB_F} \phi(\gamma, \kappa) \\
	 \text{Opt}(D) = \max_{\kappa \in \cB_F} \underline{\phi}(\kappa), \quad \underline{\phi}(\kappa):= \sup_{\gamma \in \cM} \phi(\gamma, \kappa)
\end{align}
If a saddle point $(\gamma^*, \kappa^*)$ exists, then it is a primal-dual optimal pair and 
$\text{Opt}(P)=\text{Opt}(D)$. Hence, the \emph{saddle-point gap} quantifies the accuracy of a candidate solution $(\hat{\gamma}, \hat{\kappa})$:
\begin{align*}\label{eps_sp}
	\epsilon_{sp} & = \sup_{\gamma} \phi(\gamma, \hat{\kappa}) - \inf_{\kappa} \phi(\hat{\gamma},\kappa)  \\
	              & = [\overline{\phi}(\gamma) - \text{Opt}(P)] - [\text{Opt}(D) - \underline{\phi}(\kappa)]
\end{align*}
Since $\phi$ is continuous and convex-concave, and $\cM,\cB_F$ are convex and bounded, a solution always exists.

Although more involved than the alternative convex optimization approach, this saddle-point formulation results in a smooth objective, which allows for the use of methods with $O(\frac{1}{t})$ convergence rate instead of $O(\frac{1}{\sqrt{t}})$. This, however, comes at the price of a higher cost per iteration. We analyze these opposing effects theoretically in the next section and empirically in Section~\ref{sec:expts}. Beyond these computational issues, the saddle-point formulation provides interesting interpretations of the structured optimal transport problem through the lens of minimax optimization and its well-known connections to game theory and robust optimization. 

\paragraph{Game Theoretic Interpretation.}
The minimax formulation \eqref{PolytopeForm} is a \textit{min-max strategy polytope} (MSP) game \citep{Gupta2016Solving}: a two-player zero-sum game with strategies played over polytopes with payoff function $\langle \gamma, \kappa \rangle$. In this optimal transport game, Player A (the \textit{minimizer}) chooses a transport plan $\gamma$ between $\mu$ and $\nu$, and Player B (the \textit{adversary}) chooses a cost matrix $\kappa$ from the set of \textit{admissible} costs, i.e.~those that lie on the base polytope implied by the submodular cost function $F$. After this, Player $A$ pays $\langle \gamma, \kappa \rangle$ to Player $B$. Since the game is guaranteed to have a Nash equilibrium, there is a pair of transport plan $\gamma^*$ and cost matrix $\kappa^*$ such that $\gamma^*$ is optimal for fixed cost $\kappa^*$ and vice versa.

The shape and size of the adversary's strategy polytope $\cB_F$, an $nm-1$ dimensional set in $\R^{n\times m}$, depends on the characteristics of $F$. The ``more submodular'' this function is, i.e., the earlier and sharper the marginal costs decrease, the larger is $\cB_F$. If $F$ is modular, the base polytope collapses to a single point, i.e., Player B plays a fixed strategy: a ground cost matrix $C$. The problem then reduces to $\min_{\gamma \in \cM} \langle \gamma , C \rangle$, i.e., the traditional optimal transport problem \eqref{eq:original_OT}.

\paragraph{Robust Optimization Interpretation.}
Problem \eqref{PolytopeForm} can also be viewed in the light of \emph{robust optimization} \citep{Ben-Tal2009Robust, Bertsimas2011Theory}, where uncertain parameters are treated in a worst-case scenario. Structured optimal transport could then be viewed as a transportation problem with uncertain cost matrix $\kappa$, where we aim for a solution that is robust to any fluctuation of costs within the confidence set $\cB_F$.

\subsection{Further related work} 
\label{sub:related_work}

\citet{Courty2017Optimal} propose to include structural information into the standard transportation cost by adding a group-norm regularizer. In contrast, our polyhedral approach directly modifies the linear cost function, does not need a regularization coefficient, allows to integrate a wide set of combinatorial functions, and directly leads to the saddle point connections. Our framework is also fundamentally different from known connections between multi-marginal optimal transport and submodularity \citep{bach2015submodular, carlier2003class, pass2015multi}; while that setting is separable over assignments $\gamma_{ij}$, the submodularity ranges across assignment pairs between two distributions.


\section{Solving the Optimization Problem}\label{sec:opt}

\subsection{A case for proximal methods} 
\label{sub:the_case_for_proximal_methods}

Most popular first-order optimization methods for constrained convex problems fall into two categories: conditional gradient and proximal methods. Methods in the former class, like the Frank-Wolfe algorithm, require solving linear minimization oracles (LMO) as a subroutine. In the case of \eqref{PolytopeForm}, this means solving a classic (non-regularized) optimal transport problem in each iteration, which is expensive. 

On the other hand, proximal methods require mirror map computations and projections. The choice of mirror map is crucial for the efficiency of these methods, and should take into account the geometry of the constraint set. Only if the resulting projections can be easily computed are proximal methods an attractive alternative. As we show below, for appropriately chosen mirror maps this is the case for both constraint sets in problem \eqref{PolytopeForm}. We briefly discuss all required subroutines in the next section, and present outer optimization algorithms in Section~\ref{sub:algos}. We outline the main concepts here; detailed derivations may be found in the Appendix. 

\subsection{Subroutines: projections and subgradients} 
\label{sub:proximal_maps}

\paragraph{Subgradients of $f$.} 
\label{par:subgradients_of_f}
The subdifferential of $f$ is
\begin{equation*}
	\partial f(\gamma)  =  \argmax_{\kappa \in \cB_F} \langle \kappa, \gamma \rangle.
\end{equation*}
Thus, a subgradient of $f$ is computed by a linear optimization over the base polytope, which, despite exponentially many constraints, can be solved by a simple sort via Edmonds' greedy algorithm in $O(N \log N)$ time, where $N=n\times m$ is the dimension of $\gamma$.

\paragraph{Projections on the coupling polytope.} 
\label{par:projections_into_the_matching_polytope}
If we use (negative) entropy as the mirror map in $\cM$, i.e., $\Phi_{\cM}(\gamma) := H(\gamma) = \sum_{i,j} \gamma_{ij} \ln(\gamma_{ij})$, the projection of a point $w$ onto $\cM$ is given by the KL-divergence:
\begin{equation}\label{proj_coupling}
  \hat{\gamma} = \argmin_{\gamma \in \cM} \KL{\gamma}{w}.
\end{equation}
This problem is efficiently solved by the Sinkhorn-Knopp algorithm \citep{cuturi2013sinkhorn, benamou2015iterative}. An $\epsilon$-accurate solution can be computed in $O(N\log N\epsilon^{-3})$ time \citep{altschuler2017near-linear}, but often much faster empirically \citep{cuturi2013sinkhorn}.

\paragraph{Projections on the base polytope.} 
\label{par:projections_into_the_base_polytope}
If we use $\Phi_{\cB_F}(\kappa) = \tfrac12\|\kappa\|^2$, the resulting Euclidean projection\footnote{Perhaps surprisingly, the projection onto the base polytope resulting from choosing $\Phi_{\cB_F}(\kappa) := H(\kappa)$ instead is also solved by \eqref{eq:bpp} \citep{Djolonga2015Scalable}, and hence we may implement mirror descent with either projection.} on the base polytope,
\begin{equation}\label{eq:bpp}
  \hat{\kappa} = \argmin_{\kappa \in \cB_F} \| \kappa - w \|_2^2  \; = \; \argmin_{\kappa' \in \cB_{F-w}} \|\kappa'\|_2^2 + w
\end{equation}
is equivalent to minimizing the ``shifted'' submodular function $F(S) - \sum_{i \in S}w_i$ and can be computed, for instance, via the Fujishige-Wolfe minimum norm point (MNP) algorithm \citep{Wolfe1976Finding, Fujishige2006MNP}, via parametric submodular minimization and with recent cutting-plane algorithms \citep{lee2015faster}. These generic methods are nevertheless computationally very expensive, except for small problems. But most of the functions of interest, such as the group functions defined in Section~\ref{sec:submod_funs}, have additional structure: they are of the form $F(S) = \sum_{i=1}^k F_i(S)$ (also called \emph{decomposable}), each $F_i$ with small support or ``simple'' structure. Simple means that the minimum norm point problem can be solved fast. For the functions defined in \eqref{eq:cluster_funs}, and more generally, for certain hierarchical functions \citep{hochbaum95,iwata2004network}, coverage functions \citep{stobbe10efficient} and graph cuts on lines (equivalent to Total Variation), this can be solved in $O(m \log m)$ time, where $m$ is the support size of the respective $F_i$. We provide an $O(m \log m)$ algorithm for our cluster functions in the Appendix. If the supports of the $F_i$'s are disjoint, then the base polytope is a product of polytopes $\cB_{F_i}$, and the projection can be computed for each $\cB_{F_i}$ separately in parallel. If the supports overlap, then we can still exploit decomposition structure via randomized coordinate descent \citep{Ene2017Decomposable}, operator splitting methods \citep{Jegelka2013Reflection,nishihara14linear} or others \citep{stobbe10efficient} for fast optimization.

\subsection{Optimization Algorithms}\label{sub:algos}

\subsubsection{Convex formulation} 
\label{sub:convex_optimization}

We can solve the left hand side of \eqref{PolytopeForm} using mirror descent (MDA), shown as Algorithm~\ref{alg:MDA}. The choice of entropy mirror map $\Phi(\gamma) = H(\gamma)$ means that every iteration will require a KL-projection onto the base polytope and a subgradient computation, bringing the total cost per iteration to $O(N \log N + N (\log N) \epsilon^{-3})$.  For a non-smooth, not strongly convex function like the Lov\'asz extension, MDA converges with rate $O(\frac{1}{\sqrt{t}})$.

\subsubsection{Saddle-point formulation} 
\label{par:saddle_point_formulation}

We solve the minimax formulation of problem~\eqref{PolytopeForm} via either saddle-point mirror-descent (SP-MD) or saddle-point mirror-prox (SP-MP) \citep{Juditsky2011First,Juditsky2011Second}. The latter enjoys a faster convergence rate, at the cost of doubling the per-iteration cost, requiring two projections onto each of $\cM$ and $\cB_F$. In either case, the setup is as follows. Let $\Phi_{\cM}(\gamma)$ and $\Phi_{\cB_F}(\kappa)$ be mirror maps on $\cM$ and $\cB_F$, then the mirror map for the joint variable $z = (\gamma,\kappa) \in \cZ := \cM \times \cB_F$ is $\Phi(z) = \Phi_{\cM}(\gamma) + \Phi_{\cB_F}(\kappa)$, and a first-order oracle $F$ for $\phi$ is required to obtain subgradients in $\partial \phi(z)= \{ \partial_\gamma \phi(\gamma,\kappa)\} \times \{ \partial_\kappa[-\phi(\gamma,\kappa)]\}$. Thus, both the gradient computation and projection decouple over $\kappa$ and $\gamma$, and we can use the projections described in Section \ref{sub:proximal_maps}. The final SP-MP method for solving problem \eqref{PolytopeForm} is shown as Algorithm~\ref{alg:SPMP}, while the (simpler) SP-MD can be found in the Appendix. Compared to MDA and SP-MD, the mirror prox version enjoys a better convergence rate of $O(\frac{1}{t})$. Using the fast projection method for the cluster-based functions proposed here (Eq.~\ref{fun:sum_concave}), the total cost per iteration in either SP-MD and SP-MP is $O(N (\log N) \epsilon^{-3} + 
K\log K)$, where $K$ is the size of the largest cluster. 

\paragraph{Initialization} 
\label{par:initialization}
A simple choice for $\gamma_0$ is $\mu \nu^T$. For $\kappa_0$, a random corner in the base polytope\footnote{E.g.~by evaluating $f$ for a random $w \in \R^{n\times m}$.} can be used, however, we found that initializing it as the projection of $C$ onto $\cB_F$ often results in faster convergence..


\begin{algorithm}[tb]
    \caption{MDA for Structured Optimal Transport}\label{alg:MDA}
\begin{algorithmic}
   \STATE {\bfseries Input:} Initial point $\gamma_0$ and initial step size $\eta_0$
   \WHILE{$\epsilon < tol$}
   		\STATE $g_t \gets $\textsc{Edmonds}($f,\gamma_t)$
		\STATE $\tilde{\gamma}_{t+1} \gets $\textsc{Sinkhorn}$(\gamma_t \circ \exp \{ - \eta_t g_t\})$
		\STATE $\gamma_{t+1} \gets [\sum_{s=1}^{t+1}\eta_s]^{-1}\sum_{s=1}^{t+1}\eta_s \tilde{\gamma}_s$
		\STATE $\epsilon \gets  f(\gamma_{t}) - f(\gamma_{t+1})$
		\STATE $t \gets t+1$
   \ENDWHILE
\end{algorithmic}
\end{algorithm}

\begin{algorithm}[tb]
    \caption{Saddle Point Mirror Prox for Structured Optimal Transport}\label{alg:SPMP}
\begin{algorithmic}
   \STATE {\bfseries Input:} Initial point $z^0=(\gamma_0, \kappa_0)$ and step size $\eta_0$
   \WHILE{$\epsilon_{SP} < tol$}
	    \STATE // Mirror step on true gradient
		\STATE $u_{t+1} \gets $\textsc{Sinkhorn}$(\gamma_t \circ \exp \{ - \eta_t \kappa_t\})$
		\STATE $v_{t+1} \gets $\textsc{BasePolyProject}$(\kappa_t + \eta_t \gamma_t)$
	    \STATE // Mirror step on proxy gradient		
		\STATE $\gamma_{t+1}  \gets $\textsc{Sinkhorn}$(\gamma_t \circ \exp \{ - \eta_t v_{t+1})$
		\STATE $\kappa_{t+1} \gets $\textsc{BasePolyProject}$(kappa_t + \eta_t u_{t+1})$
	    \STATE // Compute saddle point gap of current solution		
		\STATE $z^{t+1} \gets [\sum_{s=1}^{t+1}\eta_s]^{-1}\sum_{s=1}^{t+1}\eta_s(\gamma_s,\kappa_s)$
		\STATE $\epsilon_{SP} \gets $\textsc{SaddleGap}$(z^t)$
		\STATE $t \gets t+1$
   \ENDWHILE
\end{algorithmic}
\end{algorithm}


\begin{figure}[ht]
\centering
  \includegraphics[scale=0.20,trim={1.1cm 0 0 0},clip]{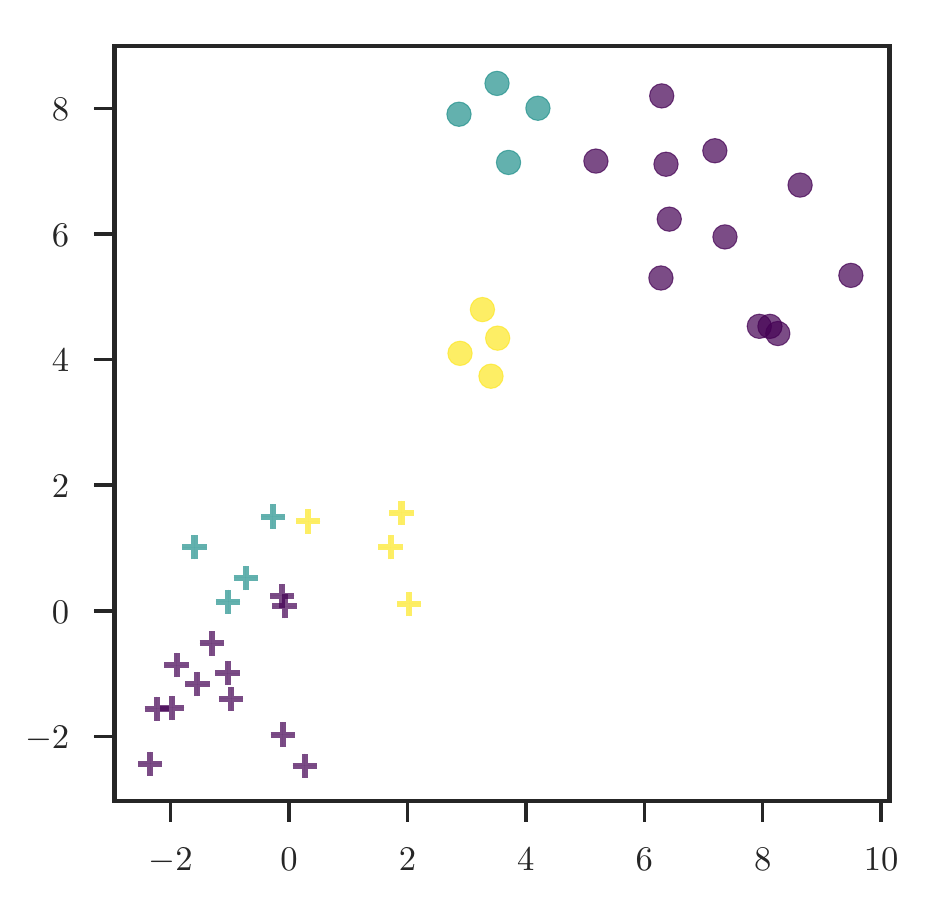}%
  \includegraphics[scale=0.4]{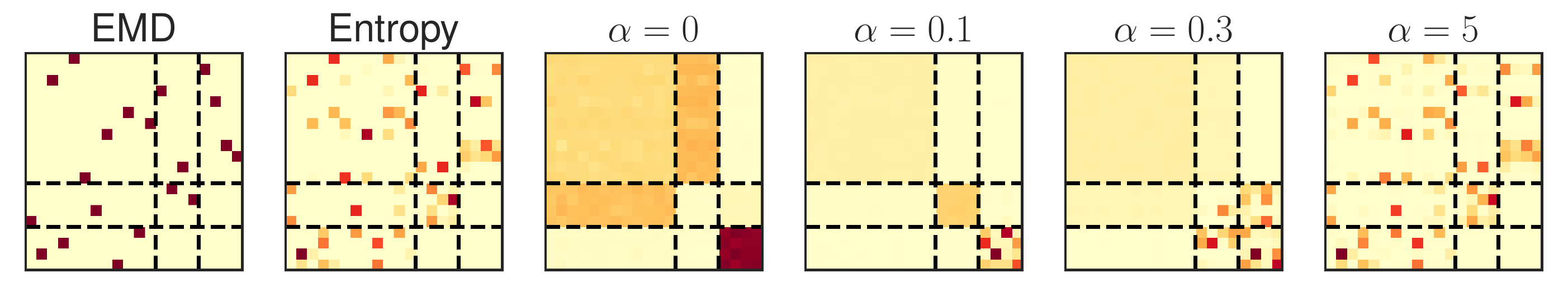}%
  \includegraphics[scale=0.25]{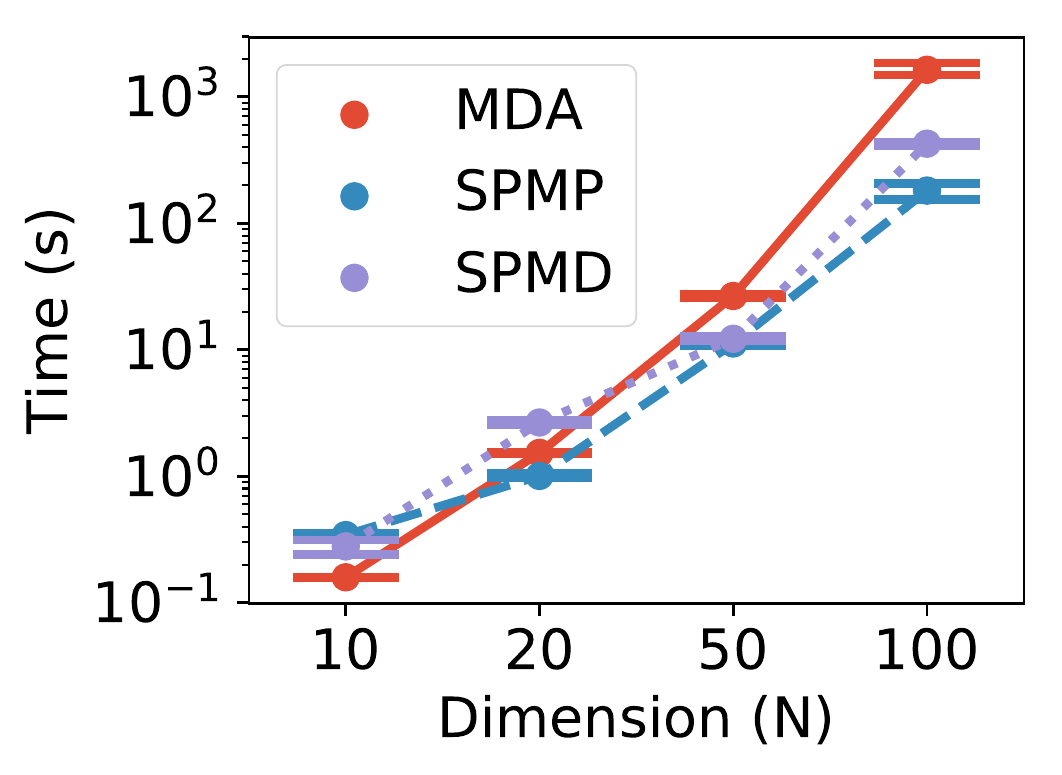}  
  \caption{Clustered point matching example. \textbf{Left:} Source and target points. \textbf{Center:} Optimal transport plans for EMD, entropy-regularized OT, and submodular optimal transport with threshold concave functions with value $\alpha$ (Eqn.~\eqref{fun:thresh_sqrt}). Dashed lines show class partitions. \textbf{Right:} Runtimes for alterative optimization methods.}\label{fig:clustmatch_gammas}
\end{figure}

\section{Experimental Results}\label{sec:expts}

Our implementation of Algorithms~\ref{alg:MDA} and \ref{alg:SPMP} uses the Python Optimal Transport library \citep{flamary2017pot} for entropic projections onto the transport polytope. For the projections onto the base polytope required by SP-MD (Alg.~\ref{alg:SPMP}), we used a tailored algorithm for decomposable functions (details in the Appendix) and RCDM \citep{Ene2015Random} when the supports are not disjoint. All experiments were run on a 2.8GHz Intel Core i7 Processor with a 16GB Memory card.  

\subsection{Clustered Point Cloud Matching}

\begin{figure}
  \includegraphics[width=\linewidth, trim={0.25cm 0.5cm 0.25cm 0.25cm}, clip]{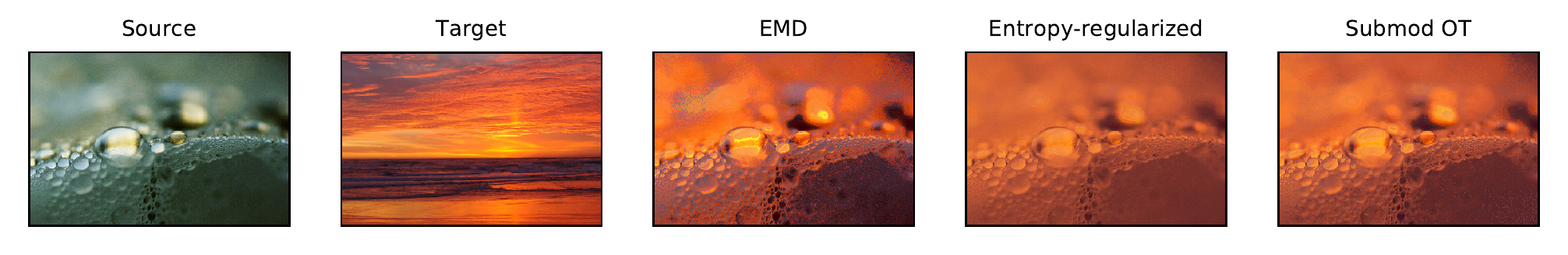}
  \caption{Color transfer with various optimal transport methods. The pixels in the source image get their color from the transported pixels in the target image. }\label{fig:color_transfer}
\end{figure}

\paragraph{Synthetic Point Clouds.} 
\label{par:point_clouds}
In our first set of experiments, we seek to understand the characteristics of the transport plans obtained with our structured optimal transport (SOT) framework. For this, we generate two point clouds in $\R^2$ with an underlying cluster structure. We use the class labels to define a sum-of-clusters function as in \eqref{eq:cluster_funs}, using square-root thresholding functions \eqref{fun:thresh_sqrt} for varying values of $\alpha$. The points and the optimal coupling matrices are shown in Figure~\ref{fig:clustmatch_gammas}. As expected, lower values of $\alpha$ enforce cluster structure more aggressively, while for larger values of $\alpha$ the cost effectively becomes modular, causing the solution to resemble those of the original transport formulations. The empirical runtimes show that SP-MP generally outperforms both SP-MD and MDA except in the very low-dimensional regime.

\paragraph{Color transfer} 
\label{par:color_transfer}
An interesting application of this matching with group information is color transfer. Here, we seek to transfer the colors of one image (the \emph{target} color scheme) into another one, the \emph{source}. To do so, we view pixels as points in RGB space, transport them using optimal transport, and assign their color to the matched pixels. Here we define partitions through superpixels obtained by segmentation \citep{felzenszwalb2004efficient}. The example in Figure~\ref{fig:color_transfer} shows that including structure in the cost function results in a coloring scheme that is more uniform that the EMD variant and sharper than the entropy-regularized one.

\begin{figure}[ht!]
	\centering
	  \includegraphics[width=0.9\linewidth, trim = {0 .3cm 0 .1cm}, clip]{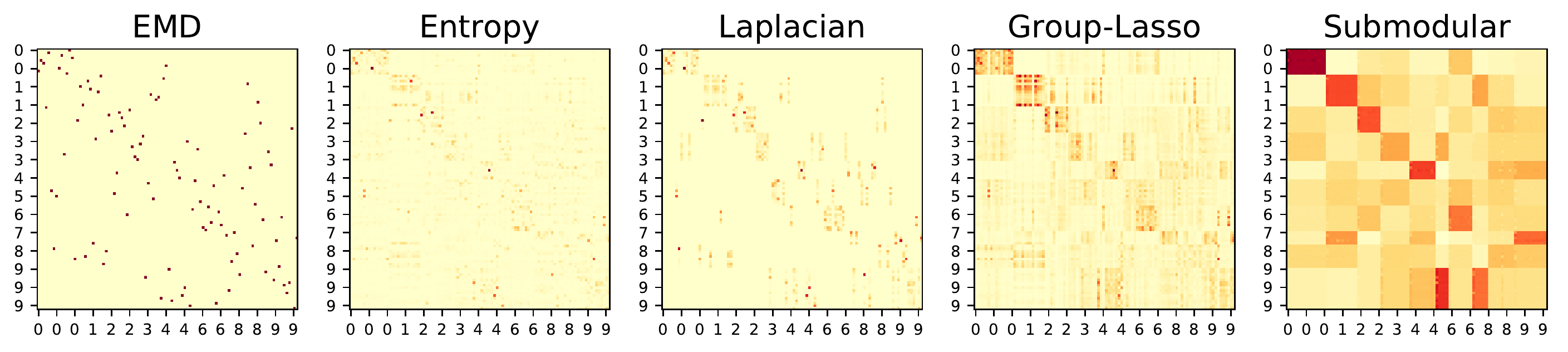}
	\caption{Optimal transport plans for the \textsc{mnist}$\rightarrow$\textsc{usps} adaptation task. Rows and columns are sorted by class.}\label{fig:DA_gammas}
\end{figure}

\subsection{Domain Adaptation}

Domain adaptation can be naturally formulated as a transportation problem: when modeling the source and target distributions via discrete samples, DOT yields an optimal transport plan $\gamma^*$ between the two samples, according to which source points can be ``transported'' to the target domain through the \emph{barycentric} mapping implicitly defined by $\gamma^*$ \citep[Chapter~7]{villani2008optimal}.

In our motivating example of domain adaptation for classification, we wish to incorporate any available class labels on either domain into the cost function, so as to encourage points of the same class to be mapped to the same region of the target space. This is seamlessly feasible via our proposed framework and the cluster functions defined before \eqref{eq:cluster_funs}. In the experiments below, we partition the source samples according to their class label, but we do not use the target labels (i.e.~every target sample forms its own cluster), so as to simulate the harder---and more realistic---unsupervised domain adaptation setting.

We test this adaptation approach on the benchmark USPS and MNIST digit classification datasets. We preprocess the data by normalizing the inputs, downscaling MNIST to the $16\times16$ size of USPS. We simulate an extreme adaptation setting where only 100 samples of each source and target domains are provided, and only source labels are available. We train a 1-NN classifier on the transported samples, and use it to predict labels on the test set (10K examples for MNIST, ~2K for USPS). We compare our method against the two class-regularized optimal transport formulations of \citet{Courty2017Optimal}: one using an $\ell_p-\ell_1$ group-sparsity norm, and the other a Laplacian regularization term. We also compare against the original formulation and the entropy-regularized version, neither of which has access to class labels. The results in Table~\ref{tab:DA_results} show that the submodular formulation achieves better accuracy in both directions of adaptation, and exhibits a much clearer block-diagonal structure in the coupling matrix (Figure~\ref{fig:DA_gammas}). We emphasize that the target labels are not used when defining the groupings of the submodular function, so this block structure is obtained solely by encouraging source points with the same label to be mapped together. Example source digits and their transported version are shown in the Appendix.

\begin{table}
	\centering
  \footnotesize
  	\begin{tabular}{ccc}
  	\toprule
  	\textbf{Method} & \textbf{MNIST}$\to$\textbf{USPS} & \textbf{USPS}$\to$\textbf{MNIST} \\
  	\midrule
	No adaption            & 41.20  & 33.10   \\
	\midrule
  	EMD 			    & 37.72  & 33.68    \\
  	Entropy 			& 55.70  & 43.64    \\
  	Laplace 			& 54.37  & 37.73   \\
  	Group-Lasso     & 57.12  & 49.49   \\
  	\textbf{Struct-OT}		    & \textbf{62.97}  & \textbf{58.34}  \\	
  	\bottomrule
  	\end{tabular}
\caption{Results on digit recognition adaptation. Values shown are prediction accuracy ($\%$).}\label{tab:DA_results}	
\end{table}

\subsection{Syntax-aware Word Mover's Distance}

The \emph{Word Mover's Distance} (WMD) is an application of optimal transport to natural language processing \citep{Kusner2015WMD}. It measures dissimilarity between strings (sentences or documents) by computing the cost of ``moving'' the words from one to the other, using a ground metric of distances between vector-space embeddings of words. The WMD, however, is syntax-agnostic, i.e., it does not take into account word ordering. That is, the cost of ``moving'' a word $u_i$ in sentence $U$ to $v_j$ in sentence $V$ depends only on their distance in the embedded space, and not on their relative positions in the two sentences. When using WMD to predict sentence similarity of long sentences with subclauses, this approach can have obvious drawbacks, like transporting words across noun-phrase boundaries.

\begin{figure}[t]
    \centering
	\includegraphics[scale=0.4]{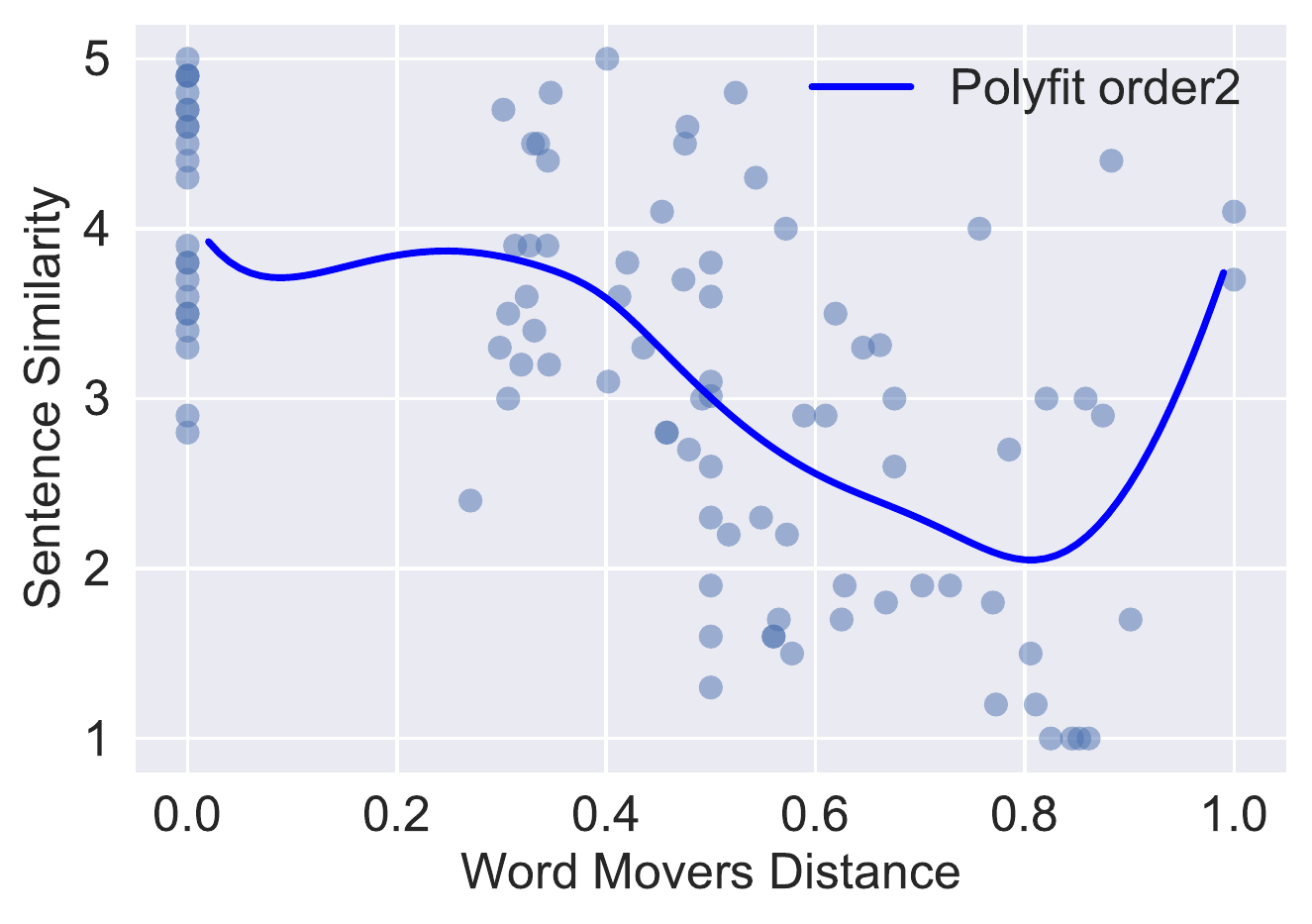}%
	\includegraphics[scale=0.4]{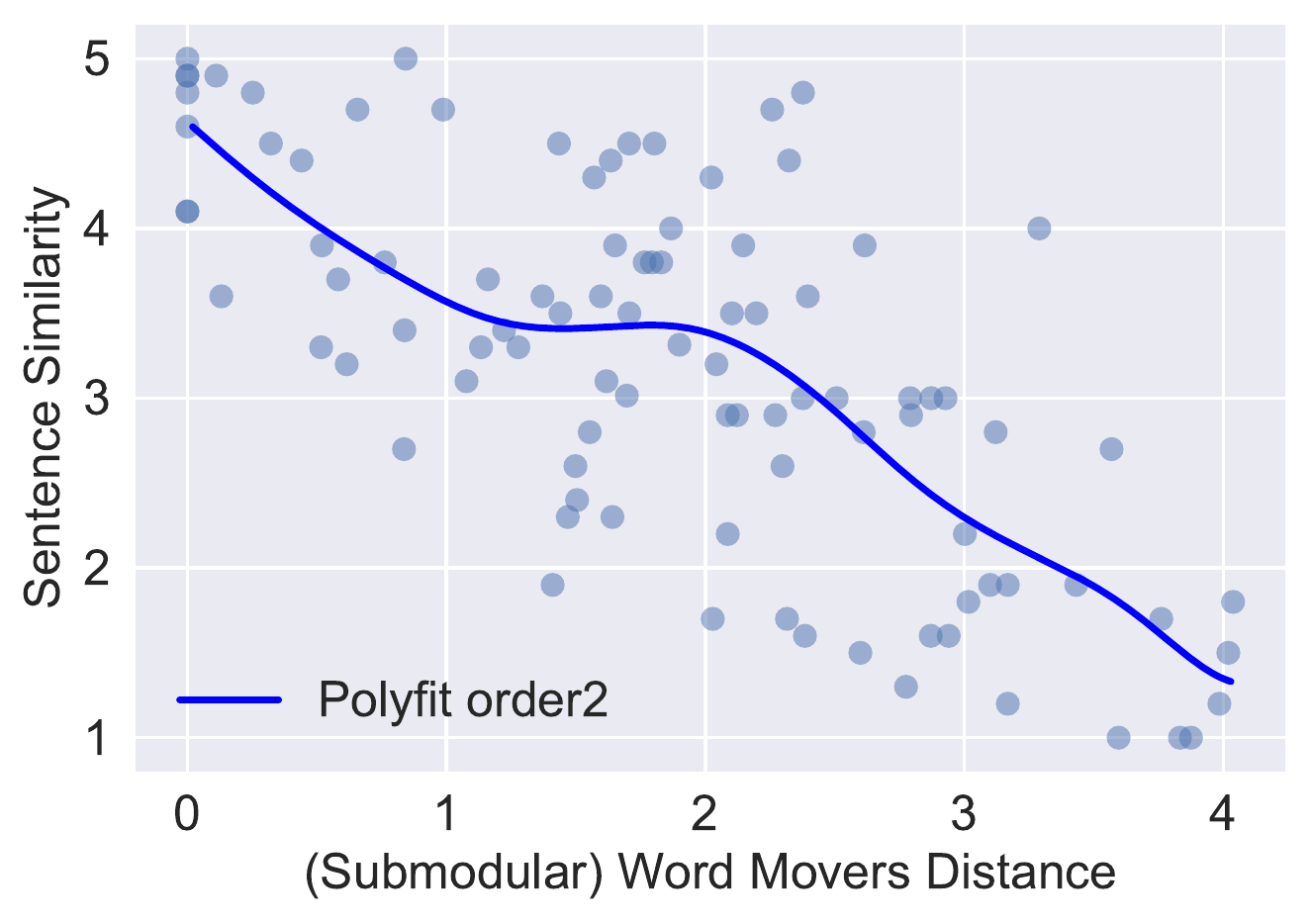}	
	\caption{Sentence similarity prediction with two classes of optimal transport distances over sentences.\label{fig:combined_plot}}
\end{figure}

We can obtain a syntax-aware alternative to WMD with a simple clustered cost function as before, where now each $n$-gram in a sentence defines a group, so we allow overlaps between the groups. With this, we are encouraging neighboring words in a sentence to be matched to neighboring words in the other. Word-to-word costs are defined as before. We compare this distance against the original WMD in a simple sentence similarity task: the SICK dataset, consisting of pairs of English sentences labeled with human-generated similarity scores. We randomly select 100 sentences from the train and test folds, we compute optimal transport distances between all training pairs, and then fit a non-parametric regression model to predict similarity scores from these distances. At test time, given a pair of sentences, we compute the distance between them and use the regression model to predict their similarity. The distances, gold similarity scores and fitted models are shown in Figure~\ref{fig:combined_plot}. The WMD model obtains a mean squared error of $0.67$ (Spearman's $\rho$ of $.71$), while our proposed syntax-aware version has a much better correlation with gold similarity scores (MSE=$0.59$, $\rho=.75$).

\section{Discussion}

We proposed a generic framework for including structural information into optimal transport problems, which are finding a growing range of applications in machine learning. While we demonstrated the utility of the framework via examples in domain adaptation, color transfer and sentence similarity, our framework can encode a variety of structures beyond these settings, since it allows arbitrary submodular functions. This choice will depend on the specifics of the problem and the efficiency with which the projections can be solved. The overall resulting convex optimization problem is efficiently solvable via mirror descent methods. For very large problems or general submodular functions, approximate or stochastic submodular optimization subroutines (if applicable) may be suitable.

In fact, the flexibility of our framework goes beyond submodularity; any convex function with bounded closed gradient maps would work as $f$. We explicitly chose submodular functions due to their favorable geometry and resulting tractability, and their ability to encode a wide range of combinatorial structures.

\bibliographystyle{abbrvnat}
\bibliography{SubdmodOT}

\pagebreak
\clearpage

%
%
\appendix

%
%
%
%

\section{The structured optimal transport is a semi-metric} 
\label{sec:the_structured_optimal_transport_is_a_metric}

We restate Lemma~\ref{lemma:semi-metric} and prove it. 

\begin{lemma}
	Suppose the ground cost $C(\cdot, \cdot)$ is a metric and that $F$ is a submodular non-decreasing function such that $F(\emptyset)=0$ and $F(\{(i,j)\}) >0 $ iff $C(x_i,y_j)>0$. Then $d_F(\mu,\nu) = \min_{\gamma \in \cM}f(\gamma)$ is a semi-metric.
\begin{proof}
Let $\mathbf{C} \in \R^{n\times m}$ be the cost matrix associated with $C$, i.e.~$\mathbf{C}_{ij} = C(x_i, y_j)$ for $i \in \{1,\dots,n\}, j\in \{1, \dots, m\}$. In addition, define $u$ and $v$ to be the vectors of probability weights of $\mu$ and $\nu$, respectively, i.e. $\mu = \sum_i^n u_i x_i$ and $\nu = \sum_j^m v_j y_j$.
	
Since $C(\cdot, \cdot)$ is a metric, $\mathbf{C}$ has only non-negative entries. Furthermore, since we assume support points are not duplicated, it has at most $n$ zero entries, and all the rest are strictly positive. This, combined with the fact that $F$ is non-decreasing, implies $F(S) \geq 0$ for every $S\subseteq V$, and therefore its Lovasz extension must also be non-negative. In particular,
\begin{equation}\label{proof:positive}
	d_F(\mu,\nu) = \min_{\gamma \in \cM}f(\gamma) \geq 0
\end{equation}
and this holds for any $\mu, \nu$.
	
Now, suppose $\mu = \nu$. Without loss of generality, we can assume supporting points are indexed such that $x_i = y_i, \medspace i \in \{1, \dots, n\}$. In addition, we must have $u = v$, so $\gamma =  \diag(u) \in \cM$. On the other hand, since $C$ is a metric $\mathbf{C}_{ii} = 0$ for every $i$, which in turn implies that for any $\kappa \in \cB_F$ and every $i$, $\kappa_{ii} \leq  F(\{i,i\}) = 0 $. By \eqref{proof:positive} and the minimax equilibrium properties, we have
\[ 0 \leq  d_F(\mu,\nu) = \langle \gamma^*, \kappa^* \rangle \leq \langle \gamma, \kappa^* \rangle \qquad \forall \gamma \in \cM \]
In particular, for $\gamma = \diag(u)$, we get 
\[ 0 \leq  d_F(\mu,\nu) \leq \sum_{i}u_i \kappa^*_{ii}  \leq 0\]
So we conclude that $d_F(\mu,\nu)=0$.
	
Conversely, suppose that $d_F(\mu,\nu)=0$, and suppose, for the sake of contradiction that $\mu \neq \nu$. This means that at least one of the following is true:
\vspace{-0.2cm}
\begin{enumerate}[itemsep=0mm]
	\item[(i)] $u \neq v$
	\item[(ii)] the support points are different, i.e.~there is no reordering of indices such that $x_i = y_i$ for every $i$.  
\end{enumerate}
If (i) is true, $\cM$ cannot permutation matrices, so in particular $\gamma^*$ has at least $n+1$ positive entries. We can thus find a $\kappa \in \cB_F$ which has positive weights in all those entries. In that case, we have $\langle \gamma^* , \hat{\kappa} \rangle > 0$, a contradiction. Now, if on the other hand (ii) is true, then $\mathbf{C}$ has strictly less than $n$ zero entries. This, by our assumptions on $F$, means that there exist $\kappa \in \cB_F$ with less than $n$ non negative entries. Any such matrix will have $\langle \gamma^*, \kappa \rangle >0$, a contradiction.

Finally, the symmetry of $d_F(\mu,\nu)$ is trivial.
\end{proof}
\end{lemma}

\section{Topological constraints in Structured Optimal Transport }

\begin{figure}[t]
	\centering
	\includegraphics[scale=0.6]{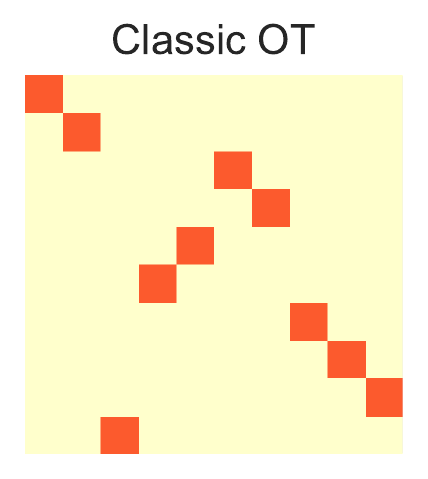}%
	\includegraphics[scale=0.28]{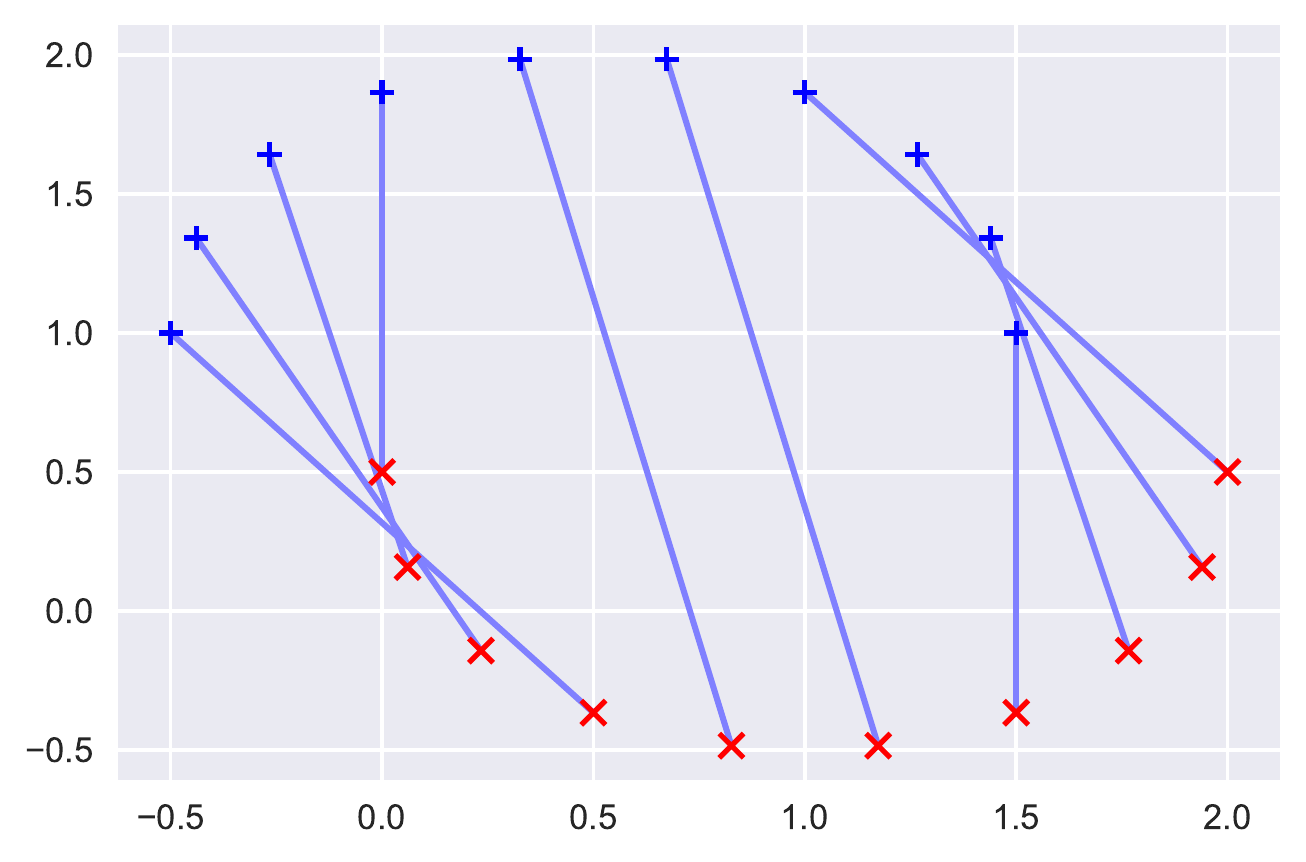}
	\includegraphics[scale=0.6]{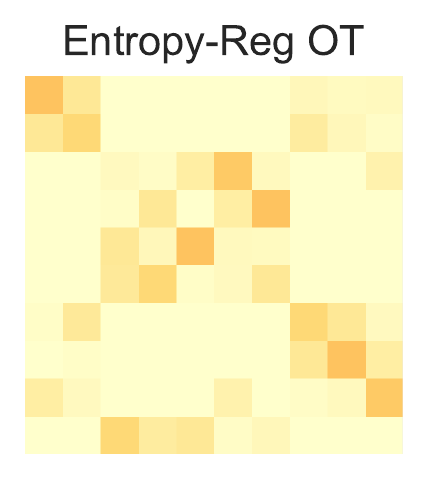}%
	\includegraphics[scale=0.3]{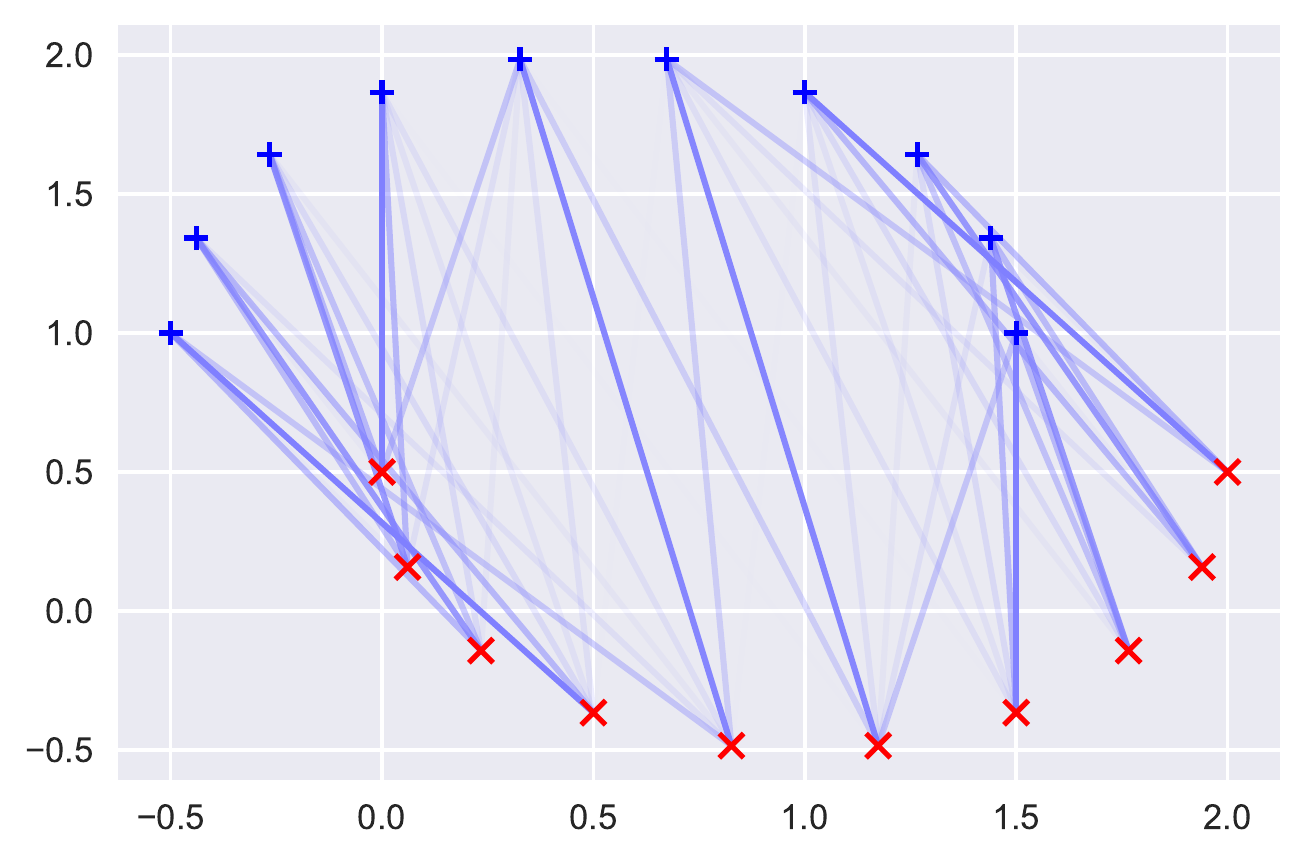}	
	\includegraphics[scale=0.6]{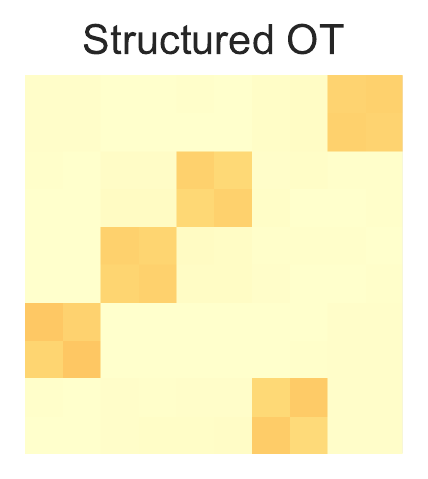}%
	\includegraphics[scale=0.3]{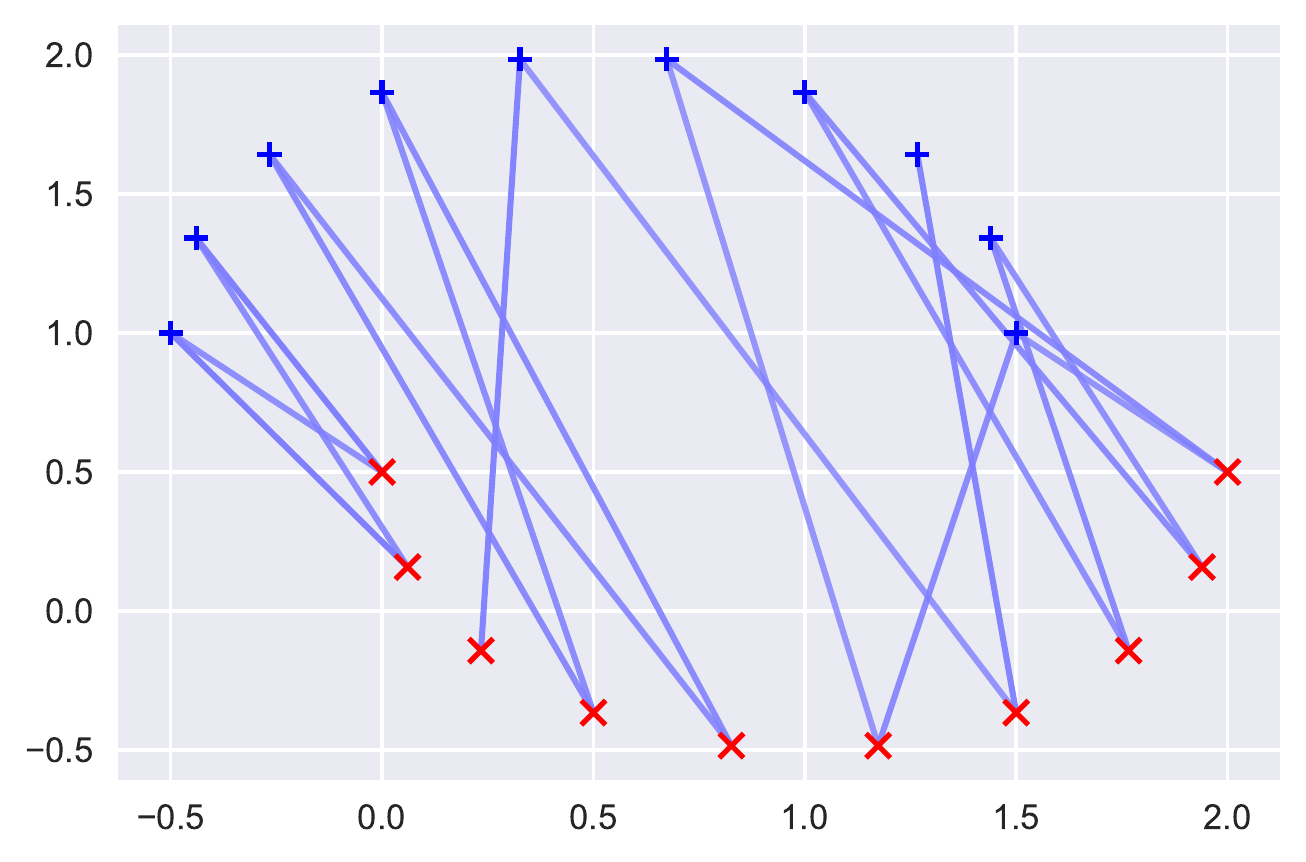}	
	\caption{Optimal transport plans and matchings for the two moons example.}\label{fig:twomoons}	
\end{figure}

Besides the settings presented above where structure arises from group labels, the framework proposed here allows us to explicitly encourage certain topological aspects of the distributions to be preserved. One such possible costraint for discrete distributions that lie on a low-dimensional manifold is to encourage neighboring points to be matched together. Such type of constraints can substantially alter the resulting transport plans, as shown in Figure~\ref{fig:twomoons} for a simple two-moons dataset. Here, the SOT solution favors neighborhood preservation over element-wise cost, resulting in a block-structured optimal coupling.

\section{The Sinkhorn-Knopp Matrix Scaling Algorithm}

\citet{cuturi2013sinkhorn} proposes to solve the entropy-regularized optimal transport problem
\begin{equation}
\argmin_{\gamma \in \cM} \langle \gamma, C \rangle - \frac{1}{\lambda}H(\gamma)
\end{equation}
with the Sinkorn-Knopp matrix scaling algorithm. Lemma 2 in \citep{cuturi2013sinkhorn}, based on Sinkhorn's Theorem \citep{sinkhorn1967diagonal}, shows that there exists a unique solution to this problem, and that it has the form 
\begin{equation*}
	\gamma_\lambda^* = \diag(u)\K\diag(v)	
\end{equation*}
where $\K$ is the entry-wise exponential of $-\frac{1}{\lambda}C$ and $u,v \in \R^d_+$. Furthermore, $u$ and $v$ can be efficiently obtained by means of Sinkhorn's fixed-point iteration, which involves updates of the form:
\begin{align*}
	u^{(n+1)} &= \mu \medspace ./ (K v^{(n)})  \\
	v^{(n+1)} &= \nu \medspace ./ (K^T u^{(n)})
\end{align*}
where, again, the division is entry-wise. The iterates $u^{(n)}$ and $v^{(n)}$ converge linearly to the true $u$ and $v$.

\section{Fast projections into submodular function base polytopes}

The problem of computing the point of minimal norm on the base polytope of a submodular function is intimately related to that of minimizing the function itself. The solutions to these two problems are related through the parametric minimization problem
\[ S_{\lambda}^* = \argmin F(S) - \lambda |S|\]
Let $\mathbf{y}^*$ be the min-norm point in $\cB_F$. We can recover the solution to the original submodular function minimization (SFM) problem, $S^*:= S^*_{\lambda=0}$ from $y^*$ as  $S^* = \{ i \suchthat y^*_i \leq 0 \}$. Conversely, we can recover $\mathbf{y}^*$ from the solutions of the parametric problem as
\[ \mathbf{y}^*_j = \max \{ \lambda \suchthat j \in S^*_{\lambda} \}  \]
Given a method for minimizing the function $F^{\lambda}: = F(S) - \lambda |S|$, one can obtain the min norm point by repeated calls to this oracle and a divide-and-conquer strategy as the one \citet{Jegelka2013Reflection} use, which runs in $O(n \log n)$ time.

Now, in our case, we are dealing with cluster functions of the form $F_i(S) = g(\sum_{i\in S} w_i)$, and in addition, we are interested in computing projections, rather than the min norm poin, i.e., $\tilde{\kappa} = \argmin_{\kappa \in \cB_{F-m}} \| \kappa \|_2^2$ for some $m$. Equivalently, we want to minimize $F_w(S) := F(S) - M(S)$, where $M$ is the modular function implied by the vector $m$.  Thus, the parametric submodular function minimization (SFM) problem we are dealing with is
\begin{align*}
	F_w^{\lambda} &= g(\sum_{i\in S} w_i) + \sum_{i \in S} m_i - \lambda |S| = g(\sum_{i\in S} w_i) + \sum_{i \in S} (m_i - \lambda) \\
	              &= \min_{\alpha \in I} c_{\alpha} + (\alpha \sum_{i \in S} w_i) + \sum_{i \in S} (m_i - \lambda) \\
				  &= \min_{u \in [0, \sum_{i \in V}w_i]} g(u) + \nabla g(u) \bigl( \sum_{i  \in S} w_i -u \bigr) + \sum_{i \in S} (m_i - \lambda)
\end{align*}
where we used the fact that any concave function can be written as the pointwise supremum of (potentially infinite) linear functions, parametrized by $\alpha$, and an interval $I$ where the valid gradients lie. Since the minimization is jointly over $S$ and $\alpha$, we can rewrite the problem as 
\begin{equation}\label{doublemin}
	\min_{\alpha} \min_S c_{\alpha} + \left ( \alpha \sum_{i \in S} w_i \right ) + \sum_{i \in S} (m_i - \lambda)
\end{equation}
As the slope $\alpha = \nabla g(u)$ shrinks, the constant $c_{\alpha} = g(u) - u\nabla g(u)$ grows. We make the following observations:
\begin{enumerate}
  \setlength\itemsep{1em}	
	\item Equation \eqref{doublemin} suggests the following strategy: (1) for each $\alpha$, find the minimizing set $S^{\alpha}$. (2) Evaluate the function above for each $S^{\alpha}$, and pick the one minimizing $F(S)$.
	\item For a fixed $\alpha$, the optimal $S^{\alpha}$ is easy to find: 
	\[ S^{\alpha} - \{i \suchthat \alpha w_i + m_i + \lambda \leq 0 \} = \{ i \suchthat \alpha \leq - (m_i + \lambda)/w_i \]
	\item Observation 2 shows that the optimal sets as $\alpha$ shrinks are nested: once an item enters the optimal set, it never leaves.
\end{enumerate}
These observtions suggest a simple sorting-based algorithm for finding the minimizer of $F(S)$, shown here as Algorithm~\ref{alg:fast_mnp}. It runs in time $O(n \log n + nT)$, where $T$ is the evaluation time of $F$ and $n$ is the size of the ground set of $F$. We emphasize that this algorithm is only valid for the concave-of-sum functions as defined in Section~\ref{sec:submod_funs}. 

\begin{algorithm}[tb]
    \caption{Fast SFM for Concave-of-Sum}\label{alg:fast_mnp}
\begin{algorithmic}
   \STATE {\bfseries Input:} Initial point $z^0=(\gamma_0, \kappa_0)$ and step size $\eta_0$
   \FOR{$i=1,\dots,n$}
   	\STATE $r_i \gets -(m_i + \lambda)/w_i$
   \ENDFOR
   \STATE $\hat{V} \gets Sort(V)$ \COMMENT{By value of $r_i$}
   \FOR{$k=1,\dots,n$}
   		\STATE $S_k \gets \{1,\dots,V(k)\}$
   \ENDFOR
   \STATE $S^* = \argmin_{S_i} F(S_i)$
   \RETURN $S^*$
\end{algorithmic}
\end{algorithm}

\begin{algorithm}[tb]
    \caption{Saddle Point Mirror Descent for Structured Optimal Transport}\label{alg:SPMD}
\begin{algorithmic}
   \STATE {\bfseries Input:} Initial point $z^0=(\gamma_0, \kappa_0)$ and step size $\eta_0$
   \WHILE{$\epsilon_{SP} < tol$}
		\STATE $\gamma_{t+1}  \gets $\textsc{Sinkhorn}$(\gamma_t \circ \exp \{ - \eta_t \kappa_t\})$
		\STATE $\kappa_{t+1}\gets $\textsc{BasePolyProject}$(\kappa_t + \eta_t \gamma_t)$
		\STATE $z^{t+1} \gets [\sum_{s=1}^{t+1}\eta_s]^{-1}\sum_{s=1}^{t+1}\eta_s(\gamma_s,\kappa_s)$
		\STATE $\epsilon_{SP} \gets $\textsc{SaddleGap}$(z^t)$
		\STATE $t \gets t+1$
   \ENDWHILE
\end{algorithmic}
\end{algorithm}

\section{Edmond's sorting algorithm}

Let $f$ be the Lovasz extension of a submodular function $F:2^V \rightarrow \R$. Then $f$ can be evaluated at $w\in \R^n$ as follows. Let $\sigma$ be a reordering of the elements of $V$ such that $w_{\sigma_1} \geq w_{\sigma_2} \geq \dots \geq w_{\sigma_n}$, and define $S_i = \{\sigma_1, \dots, \sigma_i \}$. Then 
\[ f(w) = \sum_{i =1}^n w_{\sigma_i}\bigl [ F(S_i) - F(S_{i-1}) \bigl ] \]
The computational cost in this procedure is dominated by the sorting. Now, recalling that equivalence $f(x) = \max_{y \in \cB_F} \langle y, x\rangle$, we note that this same procedure yields the maximizing $y$, setting $y_{\sigma_i} := F(S_i) - F(S_{i-1})$. It is trivial to verify that indeed $y \in \cB_F$.


\section{Derivation of Mirror Descent Steps}

We derive here the steps for the saddle-point setting The MDA version can be directly inferred from here.

Let $\cZ = \cM \times \cB_F$, and denote by $z \in \mathcal \cZ$ a pair $z = (\gamma, \kappa)$. Suppose $\Phi_{cM}, \Phi_{\cB}$ are mirror maps on $\cM$ and $\cB_F$, respectively. We define $\Phi_{\cZ}(z=(\gamma,\kappa)):= \Phi_{\cM}(\gamma) + \Phi_{\cB}(\kappa)$.

The SP-MD algorithm computes at every step:
\begin{itemize}
	\item[a)] $w_{t+1} \in D$ such that $\nabla \Phi(w_{t+1}) = \nabla \Phi(z_t) - \eta g_t$
	\item[b)] $z_{t+1} \in \argmin_{z \in \cZ} D_{\Phi}(z, w_{t+1})$ 
\end{itemize}
Note that $\Phi = ( \Phi_{\cM}, \Phi_{\cB})$, so (a) amounts to finding $w_{t+1} = (w^{\gamma}_{t+1}, w^{\kappa}_{t+1})$ such that:
\begin{align}
	\nabla \Phi_{\cM}(w^{\gamma}_{t+1}) &= \nabla \Phi_{\cM}(\gamma_{t+1}) - \eta \kappa_t \label{md:step1_M}\\ 
	\nabla \Phi_{\cB}(w^{\kappa}_{t+1}) &= \nabla \Phi_{\cB}(\kappa_{t+1}) + \eta \gamma_t \label{md:step1_K}
\end{align}
At this point, the updates take different forms depending on the mirror maps. For our choice of $\Phi_{\cM}(\gamma)=H(\gamma)$, we have $\nabla\Phi_{\cM}(\gamma) = \mathbf{1} + \log \gamma$ (where the logarithm is to be understood element-wise), so \eqref{md:step1_M} becomes: 
\begin{equation}
	\log w^{\gamma}_{t+1} = \log \gamma_{t} - \eta \kappa_{t}
\end{equation}
and thus
\[ 	w^{\gamma}_{t+1} = \gamma_{t} \cdot e^{\eta \kappa_{t}} \]
where the product and exponential are, again, element-wise. On the other hand, for the mirror map $\Phi_{\cB}(\kappa)=\frac{1}{2}\|\kappa\|_2^2$, \eqref{md:step1_K} becomes
\begin{equation}
	w^{\kappa}_{t+1} = \kappa_{t} + \eta \gamma_{t}
\end{equation}

The second step in SPMD (step (b) above) requires projecting $w_{t+1} $ and thus ($w^{\gamma}_{t+1}, w^{\kappa}_{t+1}$) into $(\cM, \cB_{F})$ according to the Bregman divergences associated with the mirror maps $\Phi_{\cM}(\gamma), \Phi_{\cB}(\kappa)$. For the entropy map, this becomes an KL-divergence projection, so we have 
\begin{equation}
	\gamma_{t+1} \in \argmin_{\gamma} \KL{\gamma}{\gamma^t \cdot e^{\eta \kappa_{t}}}
\end{equation} 
On the other hand, the divergence associated with the $\ell_2$ norm map is again an $\ell_2$ distance, so 
\begin{equation}
	\kappa_{t+1} \in \argmin_{\kappa} \| \kappa -  \kappa_{t} + \eta \gamma_{t}\|_2^2
\end{equation} 

The full SP-MD Algorithm is shown as Algorithm~\ref{alg:SPMD}. The steps of the saddle-point mirror-prox (Algorithm \ref{alg:SPMP}) can be derived analogously.

\section{Shortcomings of the Word Mover's Distance}
There are obvious limitations the WMD's purely semantic bag-of-words approach to sentence similarity, arising from ignoring the relations among words in a sentence. For example, consider the following sentences:
\begin{enumerate}[label=\emph{\alph*}),noitemsep]
 \item \emph{I left my book in the hotel}
 \item \emph{I will book my hotel }
 \item \emph{I forgot my novel in the room}
\end{enumerate}

The WMD between (a) and (b) will be less that than between (a) and (b), even though the latter two are paraphrases of each other. Even though their individual words are semantically more distant, the order in which they occur entails a very similar meaning. As contrived as this example might be, it is a good reminder that syntax and word-meaning go hand-in-hand for assessing semantic sentence similarity.

\pagebreak
\clearpage

\section{Digit transportation}
\vspace{2.5cm}
\begin{minipage}[c]{\textwidth}
	\centering
	\includegraphics[scale=0.7]{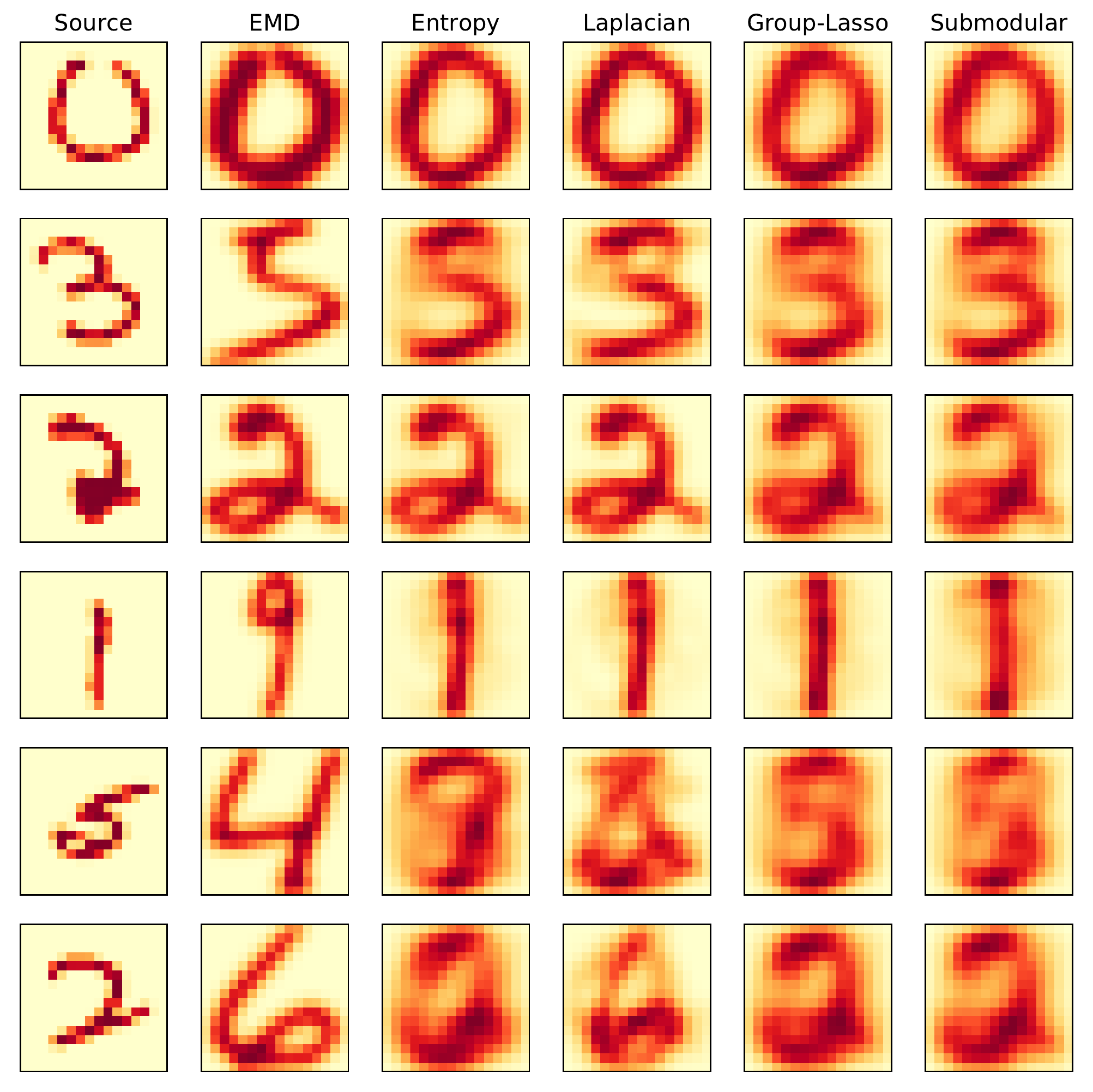}
	\captionof{figure}{Examples from the \textbf{MNIST}$\rightarrow$\textbf{USPS} domain adaptation task. The first column is the source image from \textbf{MNIST}, and the remaining columns are the result of transporting the source image with the various optimal transport maps.}	
\end{minipage}

\clearpage



\end{document}